%% file: acl_latex.tex
\pdfoutput=1

\documentclass[11pt]{article}

\usepackage[preprint]{acl}

\usepackage{times}
\usepackage{latexsym}
\usepackage{amsmath} 
\usepackage[T1]{fontenc}

\usepackage[utf8]{inputenc}

\usepackage{microtype}

\usepackage{inconsolata}

\usepackage{graphicx}
\usepackage{natbib} 
\title{Movie2Story: A framework for understanding videos and telling stories in the form of novel text}

\author{Kangning Li \\
  \texttt{kangningl25@ucla.edu} \\\And
  Zheyang Jia \\
  \texttt{u3621320@connect.hku.hk}  \\\And
  Anyu Ying \\
  \texttt{anyuy@andrew.cmu.edu} \\}

\begin{document}
\maketitle
\begin{abstract}
\input{_abstract}
\end{abstract}

\section{Introduction}

\input{1_intro}

\section{Related Works}
\input{2_rw}

\section{MSBench} 
\input{3_msbench}

\section{M2S pipeline}

\begin{figure*}
    \centering
    \includegraphics[width=1.0\linewidth]{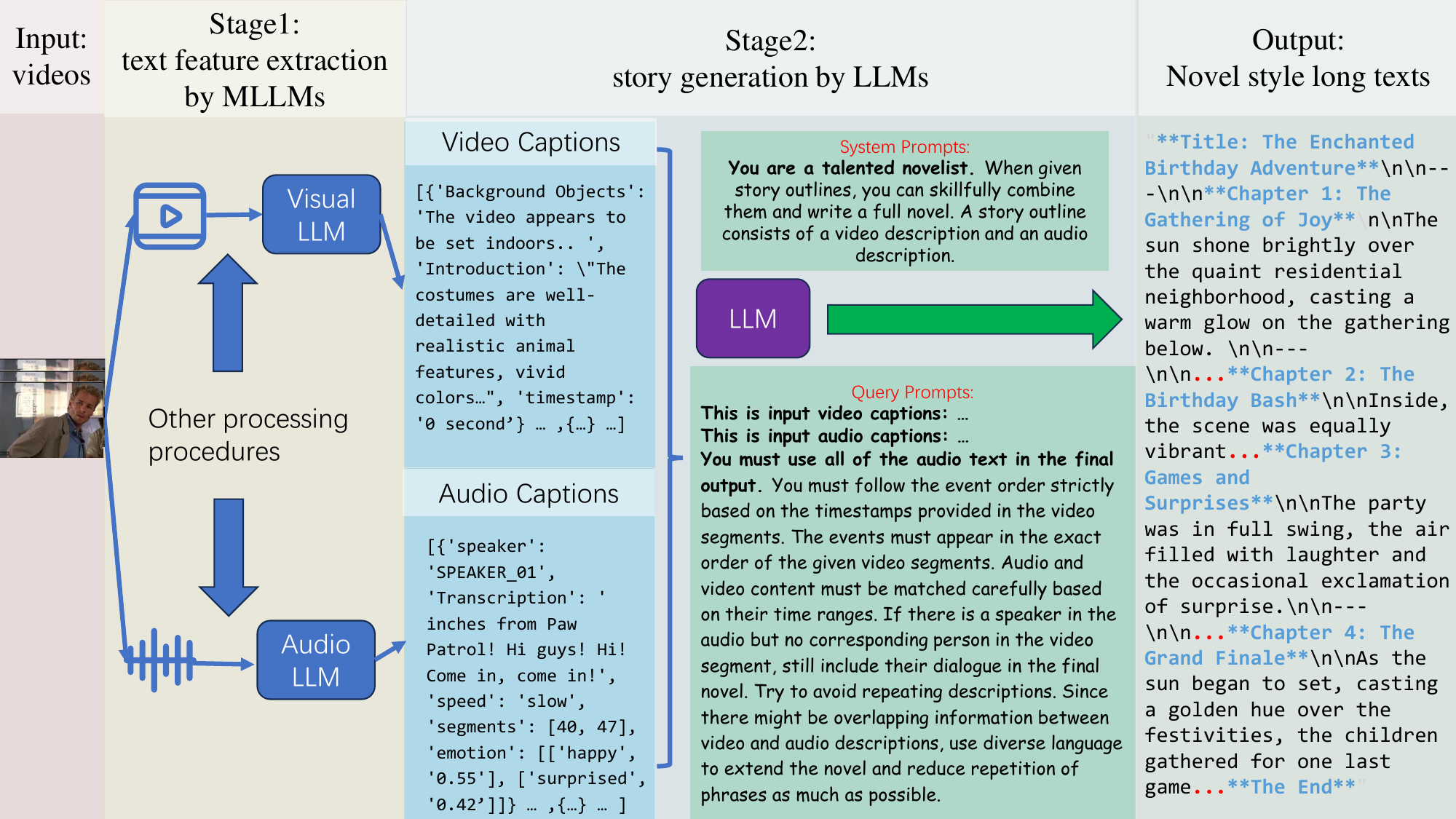}
    \caption{Workflow template}
    
    \label{fig:workflow}
\end{figure*}

\input{4_m2s}

\section{Experiment}
\input{5_exp}

\section{Conclusion}
\input{6_conclusion}


\bibliography{custom}

\appendix

\section{Appendix Tables}
\label{sec:appendix}
\input{p1}

\end{document}

%% file: _abstract.tex
In recent years, large-scale models have achieved significant advancements, accompanied by the emergence of numerous high-quality benchmarks for evaluating various aspects of their comprehension abilities. However, most existing benchmarks primarily focus on spatial understanding in static image tasks. While some benchmarks extend evaluations to temporal tasks, they fall short in assessing text generation under complex contexts involving long videos and rich auxiliary information.
To address this limitation, we propose a novel benchmark: the Multi-modal Story Generation Benchmark (MSBench), designed to evaluate text generation capabilities in scenarios enriched with auxiliary information. Our work introduces an innovative automatic dataset generation method to ensure the availability of accurate auxiliary information. On one hand, we leverage existing datasets and apply automated processes to generate new evaluation datasets, significantly reducing manual efforts. On the other hand, we refine auxiliary data through systematic filtering and utilize state-of-the-art models to ensure the fairness and accuracy of the ground-truth datasets. Our experiments reveal that current Multi-modal Large Language Models (MLLMs) perform suboptimally under the proposed evaluation metrics, highlighting significant gaps in their capabilities. To address these challenges, we propose a novel model architecture and methodology to better handle the overall process, demonstrating improvements on our benchmark.


%% file: 1_intro.tex
In recent years, many multi-modal large language models (MLLMs)\cite{alayrac2022flamingovisuallanguagemodel,zhu2023minigpt4enhancingvisionlanguageunderstanding,li2024videochatchatcentricvideounderstanding,huang2023languageneedaligningperception,Li_2020} have effectively combined video encoders\cite{touvron2023llamaopenefficientfoundation, devlin2019bertpretrainingdeepbidirectional,dosovitskiy2021imageworth16x16words} and audio encoders\cite{radford2022robustspeechrecognitionlargescale,chen2022beatsaudiopretrainingacoustic,chen2022htsathierarchicaltokensemanticaudio} to achieve strong performance in vision-based description generation and audio-based tasks. However, as MLLMs evolve, a critical challenge arises: how can we rigorously evaluate their comprehension and generation capabilities? Addressing this question is vital for assessing current models and guiding future developments.

Existing benchmarks for MLLMs primarily adopt a question-answering (QA) format\cite{yu2024mmvetevaluatinglargemultimodal,xu2023lvlmehubcomprehensiveevaluationbenchmark,xie2024funqasurprisingvideocomprehension,fu2024videommefirstevercomprehensiveevaluation,7780940,liu2024mmbenchmultimodalmodelallaround,cheng2024videollama2advancingspatialtemporal}, focusing on static image understanding. While benchmarks such as MVBench\cite{li2024mvbenchcomprehensivemultimodalvideo} extend evaluation to temporal tasks, they remain inadequate for assessing long-video comprehension and fail to incorporate rich auxiliary information such as audio. 
Although some studies, such as AV-SUPERB\cite{tseng2024avsuperbmultitaskevaluationbenchmark}, attempt to integrate audio information with visual information, they primarily focus on the evaluation of audio. Moreover, current benchmarks typically produce objective, template-like outputs, lacking the stylistic complexity of narrative storytelling. Furthermore, many benchmarks rely heavily on manual annotations, which are resource-intensive and time-consuming.

To overcome these limitations, we introduce a novel benchmark, Multi-modal Story Generation Benchmark (MSBench), which emphasizes the integration of diverse auxiliary information (e.g., audio features) to generate long-video descriptions in a narrative and information-rich style.

Our approach features an innovative data transformation pipeline tailored to enhance existing video-text datasets\cite{Li_2020}. Recognizing the brevity and limited scope of current datasets, we augment them with detailed audio and visual information. For instance, simple descriptions like "a car is driving" are enriched into nuanced narratives such as "a man sings: ‘...’, while a woman drives the car...". Auxiliary information, including audio-derived attributes such as ASR (Automatic Speech Recognition)\cite{Kheddar_2024} outputs, emotions, and sound events, is extracted using powerful open-source audio models. For visual descriptions, we leverage state-of-the-art image understanding models\cite{bai2023qwenvlversatilevisionlanguagemodel,chen2024internvlscalingvisionfoundation} to generate detailed frame-level descriptions, ensuring narrative consistency through temporal ordering.

Our benchmark offers two significant advantages. First, it drastically reduces the reliance on manual annotation by utilizing automated processes and open-source tools. Second, it introduces five novel evaluation metrics tailored to assess text quality in enriched information scenarios. These metrics span dimensions from environmental and psychological descriptions to name-entity specificity, providing a comprehensive assessment of narrative text quality in open-world, long-duration contexts.

Finally, we evaluate several state-of-the-art MLLMs\cite{li2024videochatchatcentricvideounderstanding,lin2024videollavalearningunitedvisual,wang2022internvideogeneralvideofoundation,wang2024internvideo2scalingfoundationmodels} on the MSBench benchmark, revealing substantial gaps in their performance. Models like VideoChat2\cite{li2024mvbenchcomprehensivemultimodalvideo} struggle with incorporating audio features, handling long-video contexts, and generating high-quality narrative outputs. To address these challenges, we propose a robust baseline model, M2S-LLM, specifically designed to align with our benchmark's requirements and enhance text generation quality. Experimental results demonstrate that M2S-LLM surpasses VideoChat2\cite{li2024mvbenchcomprehensivemultimodalvideo} by nearly 15\% across key metrics.

All models, datasets, and evaluation frameworks are publicly available to facilitate future research and advancements in the field.

%% file: 2_rw.tex
\subsection{MLLM Introduction}
The evolution of Large Language Models (LLMs) has accelerated research into Multi-modal Large Language Models (MLLMs)\cite{lyu2023macawllm,lee2024nowyousee,fu2024vitaopensourceinteractiveomni}, aiming to integrate diverse modalities like text, vision, and audio. Early breakthroughs, such as Flamingo\cite{alayrac2022flamingovisuallanguagemodel} and PaLM-E\cite{driess2023palmeembodiedmultimodallanguage}, showcased strong performance in multi-modal tasks by combining visual and textual modalities. Subsequent open-sourced efforts, including LLaVA\cite{liu2023visualinstructiontuning} and MiniGPT-4\cite{zhu2023minigpt4enhancingvisionlanguageunderstanding}, have expanded the scope of multi-modal instruction tuning, while VideoChat\cite{li2024videochatchatcentricvideounderstanding} and VideoChatGPT\cite{maaz2024videochatgptdetailedvideounderstanding} extended these ideas to dynamic video tasks by utilizing ChatGPT-generated annotations.

Building on these foundations, several recent works have introduced innovative approaches to tackle the complexities of video comprehension. The MMAD\cite{ye-etal-2024-mmad} model integrates video, audio, and textual information to generate concise, real-time descriptions. By leveraging pre-trained encoders and advanced fusion techniques, MMAD\cite{ye-etal-2024-mmad} enhances narrative richness and accessibility. Similarly, the framework for Distilling Vision-Language Models on Millions of Videos\cite{zhao2024distillingvisionlanguagemodelsmillions} adapts image-language models for video-language tasks, generating large-scale pseudo-labels with multi-granularity features, offering enhanced semantic and contextual understanding.

The Video-LLaMA2\cite{cheng2024videollama2advancingspatialtemporal} model aligns audio, video, and textual data into a unified vector space using the pre-trained ImageBind\cite{girdhar2023imagebindembeddingspacebind} module. Instead of training an audio-text dataset, it utilizes a video-text encoder, indirectly converting audio into text. Meanwhile, Video Storytelling\cite{Li_2020} focuses on generating textual summaries for events by selecting key frames with a reinforcement learning-based Narrator model and employing contextual embeddings through  Residual Bidirectional RNN (ResBRNN)\cite{Li_2020}, with Transformers enhancing long-range dependency modeling for richer, more cohesive stories.

Our work builds upon these advancements by addressing the limitations in existing MLLMs, particularly in handling long-duration videos, integrating auxiliary information, and generating stylistically rich narrative outputs. By combining innovative dataset enhancement techniques and robust evaluation metrics, our framework sets a new standard for multi-modal understanding and narrative-driven generation.

\subsection{Benchmark Introduction}
Traditional Vision-Language (VL) benchmarks\cite{goyal2017somethingsomethingvideodatabase,kay2017kineticshumanactionvideo, 7780940,xu2017video,xiao2021nextqanextphasequestionansweringexplaining} have predominantly focused on QA-style evaluations, targeting tasks like multi-modal retrieval and vision QA. More recent benchmarks have broadened the scope to assess integrated capabilities. For example, OwlEval\cite{ye2024mplugowlmodularizationempowerslarge} and SEED-Bench\cite{li2023seedbenchbenchmarkingmultimodalllms} introduced evaluation metrics that transcend simple model ranking, focusing on comprehensive multi-modal skills. In the video domain, benchmarks like Perception Test\cite{pătrăucean2023perceptiontestdiagnosticbenchmark} examine multi-modal video perception and reasoning, while FunQA\cite{xie2024funqasurprisingvideocomprehension} pushes reasoning boundaries using humorous or counter-intuitive content.

MVBench\cite{li2024mvbenchcomprehensivemultimodalvideo} stands out by defining over 20 tasks to evaluate MLLMs’ performance on diverse scenarios, especially temporal reasoning. The study on Synchronized Video Storytelling\cite{yang2024synchronizedvideostorytellinggenerating} demonstrates an innovative approach by utilizing additional ad keywords for generating and assessing the quality of storytelling and advertisement content, providing considerable inspiration for the field. Similarly, MMAD\cite{ye-etal-2024-mmad} highlights the evaluation of critical but non-automatically measurable elements, such as character names, enriching benchmark design. These advancements inform our framework, which aims to address gaps in evaluating narrative-driven, multi-modal tasks.

%% file: 3_msbench.tex
In this section, we delve into the details of our MSBench. Initially, we design the multi-modal story generation tasks as illustrated in Figure~\ref{fig:workflow}. Subsequently, we automatically generate caption-story pairs for evaluation, as depicted in stage 2 and outputs in Figure~\ref{fig:workflow}.

\subsection{Story Task Definition}
To craft the story tasks for MSBench, we employ a  text-to-text methodology, which involves integrating scattered text information into long texts in the form of novels.

As previously highlighted in the introduction, the majority of current MLLM benchmarks are centered around direct transformations from video to text or audio to text, often resulting in brief summary captions. Driven by this observation, we propose leveraging these existing task definitions as a foundation to systematically develop multi-modal story generation tasks.

We begin by summarizing the main tasks of video understanding and audio understanding text summaries from previous benchmarks. Then we enrich these tasks to longer stories, creating story tasks that can not be effectively solved with a single model (video or audio) and require comprehensive understanding of video and audio information. Finally, we define 5 tasks as follows. 

\textbf{video captions}. A comprehensive understanding for video information, generated by VQA model like videochat2\cite{li2024mvbenchcomprehensivemultimodalvideo}.

\textbf{Audio ASR}. Text version of the character's speech in the audio, exactly what they said in the audio.

\textbf{Audio Feature}. Emotion and speech speed summary for characters. And speaker recognition.

\textbf{Video Audio Alignment}. Matching characters in movie. 

\textbf{Timestamp matching and Story line Processing}. Timestamp alignment for video and audio information and generation of storyline.




\subsection{Automatic caption-story pairs Generation}
M2S maintains a Foundation Models Pool to store various video and audio foundation models, which own the capability to detect and annotate different features in videos. After having sufficient features, we use LLM to integrate the features in chronological order and generate a lengthy novel level story text.

M2S maintains a robust Foundation Models Pool. These models are equipped with the advanced capability to detect and annotate a myriad of features within videos. Once a sufficient array of features has been collected, we leverage a Large Language Model (LLM) to systematically integrate these features in chronological order (sorted by "timestamp" or segments as shown in Figure~\ref{fig:workflow} video/audio captions respectively), thereby generating a comprehensive, novel-level story text that is both lengthy and rich in detail.

\subsubsection{\textbf{Data Pre-processing}}
\textbf{(1) Video splitting}: The video may have a long length, and a short 20s video is enough to generate a long story text. Therefore, dividing a long video into 20s has the following advantages: 1. Increasing the amount of data 2. Unifying the data size 3. Reducing memory usage. 
\textbf{(2) FFmpeg}: A Video Frame Extraction Tool, used for extracting key frames from videos.\cite{ffmpeg}


\subsubsection{\textbf{Video-Audio Feature Information Generation.}. }

In our pipeline, feature extraction can be separate to several parts, here we list these models for extracting desired features and generating corresponding texts for our story tasks proposed in previous subsection.

\textbf{video captions}. Use Pre-trained Models(video-language model), including MMAD\cite{ye-etal-2024-mmad} and VideoChat2\cite{li2024mvbenchcomprehensivemultimodalvideo} for generating detailed descriptions and dialogue content.

\textbf{Audio ASR}. Use speech Extraction Tool, eg. Whisper\cite{openai2022whisper}  model for speech-to-text conversion, noting it generates text without capturing emotions. 

\textbf{Audio Feature}. Use Emotion Analysis Module Using BERT-like encoder networks for audio emotion analysis to extract emotional information and transform in text version, eg Emotion2Vec.\cite{ma2023emotion2vecselfsupervisedpretrainingspeech}

\textbf{Video Audio Alignment}. Character Relationship Network. Use CNNs like YOLOv8\cite{yaseen2024yolov8indepthexplorationinternal} for character recognition and  audio processing models (eg.pyannote)\cite{bredin2019pyannoteaudioneuralbuildingblocks} to split different characters. Matching them by anchors predefined by human.

\textbf{Timestamp matching and Story line Processing}.  Using audio timestamp splitting and video key frame resampling techniques to link information across different segments.

\subsubsection{\textbf{Story Generation}:}
Using Large Language Models, such as GPT-4o\cite{openai2024gpt4ocard}, used to expand and refine generated descriptions into a complete novel. Here we use llama3-3B\cite{grattafiori2024llama3herdmodels}, mistral7B\cite{jiang2023mistral7b}, Qwen7B\cite{bai2023qwentechnicalreport}
Collect all information and time sequence of video-audio text descriptions and generate a coherent story by well designed instructions and prompts.


\subsection{Prompt Design for Evaluation}
To emphasize the story line coherence of MLLMs, we craft a detailed system prompt for evaluation (see the top right of Figure~\ref{fig:workflow}). This prompt encourages MLLMs to carefully scrutinize video and audio content to generate stories, by paying attention to factors such as timestamps.



\subsection{Additional caption-stroy pair for fine-tuning LLMs}
We use GPT\cite{zhu2023minigpt4enhancingvisionlanguageunderstanding} to generate virtual data in the corresponding format to enhance data quality. For example, as shown in Figure~\ref{fig:workflow}, first provide the format and example of the corresponding data for GPT. Then let GPT generate data in the corresponding format as virtual data and add it to the dataset for LLMs to make fine adjustments. Because at this stage, we only need to adjust the long text generated in the form of a novel, focusing mainly on whether the generated story contains the information in the caption and temporal alignment, rather than whether the information in the caption is in the original video and audio. Therefore, the data generated in this way is reasonable.

%% file: 4_m2s.tex
After building our MSBench, we evaluate a number of popular video caption MLLMs in the first three rows in Table~\ref{tab:qualitative_performance_ablation}. Surprisingly, the existing MLLMs are far from satisfactory in temporal understanding. To fill the gap, we develop a robust MLLM-LLM pipeline, which is dubbed as M2S(movie to story). The basic process is the same as the way MSBench generates data. The basic process is the same as the way MSBench generates data. Here are some details of Video Audio Feature Information Generation and LLMs. The text integrated through LLM generally has a lower repetition rate and higher information score (column 2 and 3 in Table~\ref{tab:qualitative_performance_ablation}.)

\subsection{Video Feature Information Generation}
Our video processing pipeline first segments the video length, typically 20 seconds, to facilitate efficient processing and analysis. Then use open-source VLM and other models to extract video/image text descriptions from video files. The data is stored in JSON format for downstream text processing, which includes not only video content but also timestamps. Therefore, subsequent LLM can process it to obtain highly consistent text content.

\textbf{Video captions}
Our approach employs a segmented video processing strategy, which effectively addresses the challenges of handling long videos while maintaining the continuity of the narrative. Within each video segment, we focus on extracting storyline information and background context as the key components of the video segment. In terms of model selection, we conducted a comprehensive evaluation of various open-source models, including videollava\cite{lin2024videollavalearningunitedvisual} and Internvideo\cite{wang2022internvideogeneralvideofoundation} . Following a comparative analysis, we finally adopted for the open-source model Videochat2\cite{li2024mvbenchcomprehensivemultimodalvideo}  as our baseline model. By carefully designing the prompt format for VQA models, we ensured that the model could accurately extract the necessary key information, laying the foundation for in-depth analysis of the video content.

\textbf{Image Character Recognition}
Our goal is to match real-person names with video characters by recognizing and matching individuals across frames in long videos, despite challenges like varying appearances and multiple viewpoints. We use Facenet\cite{face}  for robust facial recognition and a text feature-based method with VIT\cite{dosovitskiy2021imageworth16x16words}  and LLM for short-segment matching. For global matching, we establish a dictionary of key characters with face and text features as anchors. YOLOv8\cite{yaseen2024yolov8indepthexplorationinternal} is used for pre-segmenting person bounding boxes, and we filter main characters by comparing anchor-candidate similarity. This approach combines vector and text features to effectively match individuals across scenes in long videos.

\subsection{Audio Feature Information Generation}
Our audio processing pipeline begins with the extraction of audio from video files using Python’s audio libraries. The extracted audio is then segmented into smaller, manageable chunks, facilitating efficient processing and analysis. The data is stored in JSON format for downstream text processing, which includes not only audio features and content but also timestamps. Therefore, it can be processed by subsequent LLMs to obtain highly consistent text content.

\textbf{Speaker Embedding for speaker recognition}
The core of our speaker recognition system lies in the Speaker Embedding module, powered by pyannote.audio\cite{bredin2019pyannoteaudioneuralbuildingblocks}. This module leverages deep learning models to extract unique and discriminative features from each speaker’s voice. These speaker embeddings serve as compact representations that capture the essential aspects of a speaker’s identity, facilitating accurate recognition and matching.

\textbf{Emotion Analysis}
Emotion2Vec\cite{ma2023emotion2vecselfsupervisedpretrainingspeech} is an innovative deep learning based emotion recognition tool that aims to encapsulate the essence of emotional vocabulary in a rich high-dimensional vector space. This method not only captures the subtle relationships between words, but also assigns them the emotional burden of human expression. Emotion2Vec demonstrated the power of deep learning in the field of emotion recognition, providing emotional storytelling through vector language. The model in this article is derived from the Emotion2Vec related model trained by FunASR.

\textbf{Word Speed}
This part is relatively simple, just calculate the number of words spoken by the speaker in a unit of time to obtain the speaking speed, using text descriptions such as "fast" and "slow".

\subsection{LLMs selection and LoRA Finetuning}
Different LLMs have varying abilities to integrate information and respond differently to instructions. Moreover, as a proprietary task, generic LLM may not necessarily have good results, so techniques such as LoRA may be needed to fine tune it. After extracting video and audio information, we can fine tune LoRA\cite{hu2021loralowrankadaptationlarge} for different LLMs to complete the final stage of generation and obtain stories. 

Note that the data here can be real video audio caption story pairs generated through our pipeline. As introduced in the previous section, GPT can also be used to generate virtual data in the corresponding format to enhance data quality.

%% file: 5_exp.tex
\subsection{Quantitative Analysis}

Traditional NLP metrics such as BLEU\cite{Papineni2002BLEU}, ROUGE\cite{Lin2004ROUGE}  evaluate generated text by comparing it to reference ground truth descriptions. While effective for assessing surface-level language quality, these metrics fail to capture the richness of knowledge beyond the visual domain, such as audio-related information, character details, or contextual nuances.

For instance, while a visual scene may depict a list of ingredients, a narration could highlight only a subset of those ingredients or incorporate additional elements like speaker emotions or environmental sounds, which are critical for generating a more comprehensive story. Since traditional metrics rely solely on visual references, they do not account for these extra layers of information.

To address this gap, we propose a set of reference-free metrics: \textbf{Language Fluency}, \textbf{Key-knowledge Relevance}. These metrics evaluate fluency, knowledge integration (including both visual and audio elements), and narrative alignment, providing a more accurate assessment of the generated stories and better reflecting the advantages of our multi-modal framework.

\textbf{Language Fluency:} Fluency is used to evaluate the coherence and sentence repetition rate of generated explanatory texts. To assess language fluency, we utilize a keyword-triplet method that evaluates fluency at three levels of granularity: (1) within a single sentence, (2) between different sentences, and (3) the repetition of key-knowledge triplets relative to the overall story. Repetition of triplets within the story is considered a sign of unnatural phrasing and redundancy, which negatively impacts fluency. A higher fluency score is assigned when the text flows smoothly, with minimal repetition and a coherent structure. We have made modifications to the formulas in the original paper\cite{yang2024synchronizedvideostorytellinggenerating}.Fluency and repetition are opposite. In the chart, we use repetition to represent the fluency of language, with lower repetition indicating smoother language. Then repetition can be represented by \textbf{Intra-story Repetition (ISR):}
$ 
\text{ISR}(S) = \frac{1}{n(n-1)} \sum_{i \neq j} \frac{\left| W(s_i) \cap W(s_j) \right|}{\left| W(s_i) \right|}
$
where \( n \) is the number of explanatory texts, and \( W(s_i) \) represents the vocabulary set of explanatory text \( s_i \). The formula calculates the repetition rate of vocabulary within the explanatory texts, which should be minimized to ensure diversity and coherence in the text. 

\textbf{Key-knowledge Relevance:} For evaluating the relevance of key knowledge in the generated story, we define a knowledge base that includes both visual and audio information. In contrast to traditional models that focus solely on visual knowledge, our approach incorporates additional audio-related knowledge, such as ASR (Automatic Speech Recognition), environmental sound events, and other audio features. This expanded knowledge base allows us to assess the relevance and diversity of knowledge within the story. Specifically, we use two evaluation criteria:
- \textbf{Information Similarity (InfoSim)}\cite{yang2024synchronizedvideostorytellinggenerating}: Measures the alignment between the knowledge points in the story and the knowledge repository. A higher similarity score indicates that the generated story effectively incorporates relevant knowledge.
- \textbf{Information Diversity (InfoDiverse)}\cite{yang2024synchronizedvideostorytellinggenerating}: Evaluates the breadth of the knowledge used in the story. A higher diversity score indicates that the story incorporates a wide range of knowledge points, avoiding over-reliance on a small subset of information.
The formula here is quite different from the original text. We have changed the normalization for time to the normalization for the number of keywords. That is to say, InfoDiverse calculates the proportion of keywords in the generated data that match the keywords in the reference data, ranging from 0 to 1.

\begin{align}
\text{InfoSim} &= \frac{1}{2|s_i|}\sum_{s_i}\left(\max_{k\in K} f_k^T f_{s_i} \right. \notag\\
&\quad \left.+\frac{1}{|W(s_i)|}\sum_{w\in W(s_i)}\max_{k\in K} f_k^T f_w\right)
\end{align}

\begin{align}
\text{InfoDiverse} &= \frac{1}{|K|}\left|\bigcup_{s_i}\{k_t\in K \mid \max_{w\in W(s_i)\cup S_i} \right. \notag\\
&\quad \left. f_{k_t}^T f_w > 0.9\}\right|
\end{align}

where $W(s_i)$ represents all words in sentence $s_i$. $f_{s_i}$, $f_k$, and $f_w$ refer to the normalized embeddings of sentence $s_i$, knowledge point $k$, and segmented word $w$, respectively.

A well-rounded story should demonstrate both high information similarity and diversity, ensuring that relevant visual and audio knowledge are integrated in a meaningful way.


These evaluation metrics provide a comprehensive framework for assessing the quality of generated story descriptions, ensuring that the generated narratives meet high standards in terms of fluency, knowledge relevance, and visual coherence.

\subsection{Qualitative Analysis}
\subsubsection{GPT Assistant Metrics}

In this section, we introduce a set of key evaluation metrics designed to assess the quality of generated story descriptions using GPT and prompt engineering. By defining these metrics, we aim to leverage GPT as a rigorous evaluator of novel-like text, assigning scores based on specific story elements. These metrics are particularly useful for assessing narrative quality in terms of various aspects such as environmental details, emotional expression, language proficiency, character portrayal, and storyline coherence. 

The following metrics are defined:

\textbf{Environment Description Score:} This score evaluates how well the generated text sets the scene and conveys the environment. Key factors include the richness of the setting, sensory detail, and atmosphere. A high score indicates a vivid and immersive portrayal of the environment.

\textbf{Emotional Description Score:} This metric assesses the depth and authenticity of emotional expression in the story. It measures how well the text captures and conveys the emotional states of characters, as well as the emotional tone of the narrative.

\textbf{Language Description Score:} This score focuses on the clarity, fluency, and grammatical quality of the text. It evaluates how well the story is written, including sentence structure, vocabulary choice, and coherence. A higher score reflects a more polished and readable narrative.

\textbf{Character Description Score:} This metric evaluates how effectively characters are described in terms of appearance, personality, and behavior. It gauges whether the character descriptions are detailed, nuanced, and consistent throughout the narrative.

 \textbf{Structural and Rationality Coherence Index (SRCI) Score:} This metric quantifies the narrative coherence and the logical progression of the storyline. It gauges the effectiveness of plot construction, the lucidity of the story’s trajectory, and the seamless incorporation of essential storytelling components such as conflict, resolution, and character progression. The SRCI reflects the story’s ability to maintain a consistent and engaging structure that resonates with the audience’s expectations of narrative rationality.”

 \textbf{Storyline Accuracy with Canonical Outline Reference (SACOR) Score:} This score measures the accuracy of the generated story against a predefined canonical storyline. It assesses how closely the generated narrative aligns with the key plot points, character behaviors, and thematic elements of the original story outline. The SACOR score is indicative of the model’s fidelity to the source material and its ability to reproduce the intended narrative arc with precision.”

For a more comprehensive evaluation, these scores are tabulated and compared across different story generations. The results can be found in Table~\ref{tab:qualitative_performance}, which summarizes the performance of our framework on each metric.

In Table~\ref{tab:qualitative_performance}, the scores represent the average evaluation across multiple generated stories, providing a comprehensive view of the performance of our framework compared to baseline methods.

\subsubsection{Manual Metrics}
Table~\ref{tab:qualitative_performance} shows the results of applying our M2S framework for generating detailed story descriptions from several movies. The comparison between our framework and baseline methods demonstrates the clear advantages of our approach.\\
More specifically, our framework excels in producing more comprehensive and accurate story descriptions. By leveraging the Audio module, we incorporate key information such as ASR (Automatic Speech Recognition) output, speaker identity, and speaking rate to further enhance the generated descriptions. Additionally, the combination of the Actor-Matching module, Storyline-Segment module, Audio module, and Vision module enables our framework to generate rich, long-form narrative descriptions that go beyond traditional subtitles, offering a novel way of understanding movies. This integrated approach results in more detailed and accurate descriptions, which could serve as an independent means of movie comprehension, rather than merely being a supplementary subtitle track.

%% file: 6_conclusion.tex
In this paper, we introduced MSBench, a comprehensive benchmark designed to evaluate the multi-modal story generation capabilities of MLLMs. We also proposed a robust MLLM pipeline, M2S, which outperforms leading models on the MSBench benchmark. Our extensive analyses provide valuable insights into the design of MLLMs for multi-modal story generation, particularly in scenarios enriched with additional information. Despite these advancements, our current approach has some limitations. We aim to address these in future work to enhance the evaluation framework and further refine the performance of MLLMs in complex, information-rich environments.


%% file: p1.tex
\begin{table*}
\centering
\scalebox{0.8}{
\begin{tabular}{cccccccccc}
\hline
Model & ISR & Info\_sim & Info\_diverse & SACOR & Rouge-1 & Rouge-2 & Rouge\_L& BLEU-4 & ... \\ \hline
Internvideo2 & 15.1 &  &  &   & &  &  &  \\
VideoLlava  & 42.7 &  &  &   & &  &  &  \\
Videochat2& 8.1 & 38.0 & 55.0 &   & &  &  &  \\
\hline
Internvideo2 + GPT4o & 4.0 & 26.5 & 18.6 & 59.4  & 39.6 & 10.2 & 18.6 & 2.7 \\
VideoLlava + GPT4o & 5.6 & 25.0 & 20.7 & 61.1  & 40.9 & 10.5 & 19.0 & 3.2 \\
Videochat2 + GPT4o  & 4.5 & 27.2 & 18.3 & 61.2  & 40.1 & 10.1 & 18.8 & 2.9  \\
\hline
Internvideo2 + A-GPT4o & 5.6 & 25.5 & 17.6 & 67.0  &   &   &  &   \\
VideoLlava + A-GPT4o & 6.0 & 25.2 & 21.9 & 65.3 &  &   &  &   \\
Videochat2 + A-GPT4o & 12.4 & 27.3 & 53.6 & 52.8 &  &   &  &   \\
\hline
\end{tabular}
}
\caption{Performance of various models with GPT4o as LLM on Quantitative metrics}
\label{tab:quantitative_performance_ablation}
\end{table*}

\begin{table*}
\centering
\scalebox{0.8}{
\begin{tabular}{ccccccc|c}
\hline
Model & Environment & Character & Emotion & Language  & SRCI  & Overall GPT & Human \\
\hline
Internvideo2& 2.494 & 1.525 & 0.608 & 1.825 & 1.421 & 1.575  \\
VideoLlava   & 2.170
 & 1.155
 & 0.213
 & 1.539
 & 1.118
 & 1.239 \\
Videochat2  & 3.086
 & 1.878
 & 0.921
 & 2.243
 & 1.809
 & 1.987  \\
\hline
Internvideo2 + GPT4o & 3.766
 & 2.675
 & 1.792
 & 3.555
 & 2.717
 & 2.901 \\
VideoLlava + GPT4o  & 3.592
 & 2.586
 & 1.667
 & 3.438
 & 2.594
 & 2.775 \\
Videochat2 + GPT4o & 3.873
 & 2.651
 & 1.738
 & 3.550
 & 2.727
 & 2.908 \\
\hline
Internvideo2 + A-GPT4o& 3.581
 & 2.574
 & 1.659
 & 3.433
 & 2.585
 & 2.767 \\
VideoLlava + A-GPT4o& 3.427
 & 2.491
 & 1.607
 & 3.195
 & 2.485
 & 2.641  \\
Videochat2 + A-GPT4o & 3.492
 & 2.568
 & 1.673
 & 3.090
 & 2.625
 & 2.689  \\
\hline
\end{tabular}
}
\caption{Performance of various models with GPT4o as LLM on Qualitative metrics}
\label{tab:qualitative_performance_ablation}
\end{table*}

\begin{table*}
\centering
\scalebox{0.8}{
\begin{tabular}{ccccccccccc}
\hline
Model & ISR & Info\_sim & Info\_diverse& SACOR &Rouge-1 & Rouge-2 & Rouge\_L& BLEU-4 & ...\\
\hline
GPT4o& 12.4 & 27.3 &\textbf{53.6} & 52.8  & -  & - & - & - \\
GPT3.5 & 13.5 & 26.3 & 25.5 & 54.0 & 38.2 & 8.6 & 17.6 & 1.9  \\
Doubao& 5.4 & 27.7 & 16.0 & 56.8 & 32.6 & 7.0 & 15.7 & 0.8 \\
QWen & 3.4 & 33.2 &46.3  & \textbf{60.1}  & 46.2 & 17.8 & 22.3 & 5.8 \\
QWen-7B-instruct& 9.8 & 35.3 & 50.8 & 58.3 &21.4  &  \textbf{34.8} & 16.5 & \textbf{27.3}  \\
mistral-instruct-7B-v1.0 & 8.1 &  & 8.3 &7.5  & 3.5 &  4.7 & 1.8 &  6.3 \\
llama3-3B & 22.1 & - & 7.9 & 16.6  &  8.6 &  21. & 24.6 &  8.2 \\
lora\_QWen-7B-instruct& 7.6 & 33.1 & 50.8 & 58.2 & \textbf{52.4} & 21.7 & \textbf{25.1} & 9.0  \\
lora\_mistral-instruct-7B-v1.0 & 7.3 & 33.0 & 50.1 & 58.0 & 52.2 &  21.7 & 25.0 &  8.9 \\
lora\_llama3-3B & 8.5 & 33.0 & 52.4 &  58.8 &  52.2 &  21.2 & 24.6 &  8.7 \\
\hline
\end{tabular}
}
\caption{Performance of various LLMs on Quantitative metrics}
\label{tab:quantitative_performance}
\end{table*}

\begin{table*}
\centering
\scalebox{0.8}{
\begin{tabular}{ccccccc|c}
\hline
Model & Environment & Character & Emotion & Language & SRCI & Overall GPT &  Human \\
\hline
GPT4o & 3.492 & \textbf{2.558} & \textbf{1.673}  & 3.090  &  \textbf{2.625} & \textbf{2.689} & -   \\
GPT3.5 & 2.831 & 1.809 & 1.077 & 2.299 & 1.815 & 1.966 & -   \\
Doubao & 2.998 & 1.906  & 1.025  & 2.233 & 1.878  & 2.008 & -  \\
QWen & 3.491 & 2.356 & 1.567 & 2.994  & 2.376 & 2.557 &  -  \\
QWen-7B-instruct & 2.826 & 1.884 & 1.125 & 2.224  & 1.805 & 1.973 &  -  \\
mistral-instruct-7B-v1.0 & 0.360 & 0.229  & 0.124  & 0.297   & 0.213 & 0.245  &  -  \\
llama3-3B & 0.294 & 0.187 & 0.084 & 0.261  & 0.134 & 0.192 & -  \\
lora\_QWen-7B-instruct& 3.606 & 2.468 & 1.611 & \textbf{3.170}  & 2.504 & 2.672 &  -  \\
lora\_mistral-instruct-7B-v1.0 & \textbf{3.623} & 2.494 & 1.625 & 3.159  & 2.527 & 2.686 &  -  \\
lora\_llama3-3B & 3.560 & 2.431 & 1.577 &  3.106 & 2.457 & 2.626 &  -  \\
\hline
\end{tabular}
}
\caption{Performance of various LLMs on Qualitative metrics}
\label{tab:qualitative_performance}
\end{table*}




%% file: acl_latex.bbl
\begin{thebibliography}{56}
\providecommand{\natexlab}[1]{#1}

\bibitem[{Alayrac et~al.(2022)Alayrac, Donahue, Luc, Miech, Barr, Hasson, Lenc, Mensch, Millican, Reynolds, Ring, Rutherford, Cabi, Han, Gong, Samangooei, Monteiro, Menick, Borgeaud, Brock, Nematzadeh, Sharifzadeh, Binkowski, Barreira, Vinyals, Zisserman, and Simonyan}]{alayrac2022flamingovisuallanguagemodel}
Jean-Baptiste Alayrac, Jeff Donahue, Pauline Luc, Antoine Miech, Iain Barr, Yana Hasson, Karel Lenc, Arthur Mensch, Katie Millican, Malcolm Reynolds, Roman Ring, Eliza Rutherford, Serkan Cabi, Tengda Han, Zhitao Gong, Sina Samangooei, Marianne Monteiro, Jacob Menick, Sebastian Borgeaud, Andrew Brock, Aida Nematzadeh, Sahand Sharifzadeh, Mikolaj Binkowski, Ricardo Barreira, Oriol Vinyals, Andrew Zisserman, and Karen Simonyan. 2022.
\newblock \href {https://arxiv.org/abs/2204.14198} {Flamingo: a visual language model for few-shot learning}.
\newblock \emph{Preprint}, arXiv:2204.14198.

\bibitem[{Bai et~al.(2023)Bai, Bai, Yang, Wang, Tan, Wang, Lin, Zhou, and Zhou}]{bai2023qwenvlversatilevisionlanguagemodel}
Jinze Bai, Shuai Bai, Shusheng Yang, Shijie Wang, Sinan Tan, Peng Wang, Junyang Lin, Chang Zhou, and Jingren Zhou. 2023.
\newblock \href {https://arxiv.org/abs/2308.12966} {Qwen-vl: A versatile vision-language model for understanding, localization, text reading, and beyond}.
\newblock \emph{Preprint}, arXiv:2308.12966.

\bibitem[{Chen et~al.(2022{\natexlab{a}})Chen, Du, Zhu, Ma, Berg-Kirkpatrick, and Dubnov}]{chen2022htsathierarchicaltokensemanticaudio}
Ke~Chen, Xingjian Du, Bilei Zhu, Zejun Ma, Taylor Berg-Kirkpatrick, and Shlomo Dubnov. 2022{\natexlab{a}}.
\newblock \href {https://arxiv.org/abs/2202.00874} {Hts-at: A hierarchical token-semantic audio transformer for sound classification and detection}.
\newblock \emph{Preprint}, arXiv:2202.00874.

\bibitem[{Chen et~al.(2022{\natexlab{b}})Chen, Wu, Wang, Liu, Tompkins, Chen, and Wei}]{chen2022beatsaudiopretrainingacoustic}
Sanyuan Chen, Yu~Wu, Chengyi Wang, Shujie Liu, Daniel Tompkins, Zhuo Chen, and Furu Wei. 2022{\natexlab{b}}.
\newblock \href {https://arxiv.org/abs/2212.09058} {Beats: Audio pre-training with acoustic tokenizers}.
\newblock \emph{Preprint}, arXiv:2212.09058.

\bibitem[{Chen et~al.(2024)Chen, Wu, Wang, Su, Chen, Xing, Zhong, Zhang, Zhu, Lu, Li, Luo, Lu, Qiao, and Dai}]{chen2024internvlscalingvisionfoundation}
Zhe Chen, Jiannan Wu, Wenhai Wang, Weijie Su, Guo Chen, Sen Xing, Muyan Zhong, Qinglong Zhang, Xizhou Zhu, Lewei Lu, Bin Li, Ping Luo, Tong Lu, Yu~Qiao, and Jifeng Dai. 2024.
\newblock \href {https://arxiv.org/abs/2312.14238} {Internvl: Scaling up vision foundation models and aligning for generic visual-linguistic tasks}.
\newblock \emph{Preprint}, arXiv:2312.14238.

\bibitem[{Cheng et~al.(2024)Cheng, Leng, Zhang, Xin, Li, Chen, Zhu, Zhang, Luo, Zhao, and Bing}]{cheng2024videollama2advancingspatialtemporal}
Zesen Cheng, Sicong Leng, Hang Zhang, Yifei Xin, Xin Li, Guanzheng Chen, Yongxin Zhu, Wenqi Zhang, Ziyang Luo, Deli Zhao, and Lidong Bing. 2024.
\newblock \href {https://arxiv.org/abs/2406.07476} {Videollama 2: Advancing spatial-temporal modeling and audio understanding in video-llms}.
\newblock \emph{Preprint}, arXiv:2406.07476.

\bibitem[{Devlin et~al.(2019)Devlin, Chang, Lee, and Toutanova}]{devlin2019bertpretrainingdeepbidirectional}
Jacob Devlin, Ming-Wei Chang, Kenton Lee, and Kristina Toutanova. 2019.
\newblock \href {https://arxiv.org/abs/1810.04805} {Bert: Pre-training of deep bidirectional transformers for language understanding}.
\newblock \emph{Preprint}, arXiv:1810.04805.

\bibitem[{Dosovitskiy et~al.(2021)Dosovitskiy, Beyer, Kolesnikov, Weissenborn, Zhai, Unterthiner, Dehghani, Minderer, Heigold, Gelly, Uszkoreit, and Houlsby}]{dosovitskiy2021imageworth16x16words}
Alexey Dosovitskiy, Lucas Beyer, Alexander Kolesnikov, Dirk Weissenborn, Xiaohua Zhai, Thomas Unterthiner, Mostafa Dehghani, Matthias Minderer, Georg Heigold, Sylvain Gelly, Jakob Uszkoreit, and Neil Houlsby. 2021.
\newblock \href {https://arxiv.org/abs/2010.11929} {An image is worth 16x16 words: Transformers for image recognition at scale}.
\newblock \emph{Preprint}, arXiv:2010.11929.

\bibitem[{Driess et~al.(2023)Driess, Xia, Sajjadi, Lynch, Chowdhery, Ichter, Wahid, Tompson, Vuong, Yu, Huang, Chebotar, Sermanet, Duckworth, Levine, Vanhoucke, Hausman, Toussaint, Greff, Zeng, Mordatch, and Florence}]{driess2023palmeembodiedmultimodallanguage}
Danny Driess, Fei Xia, Mehdi S.~M. Sajjadi, Corey Lynch, Aakanksha Chowdhery, Brian Ichter, Ayzaan Wahid, Jonathan Tompson, Quan Vuong, Tianhe Yu, Wenlong Huang, Yevgen Chebotar, Pierre Sermanet, Daniel Duckworth, Sergey Levine, Vincent Vanhoucke, Karol Hausman, Marc Toussaint, Klaus Greff, Andy Zeng, Igor Mordatch, and Pete Florence. 2023.
\newblock \href {https://arxiv.org/abs/2303.03378} {Palm-e: An embodied multimodal language model}.
\newblock \emph{Preprint}, arXiv:2303.03378.

\bibitem[{et~al.(2024{\natexlab{a}})}]{grattafiori2024llama3herdmodels}
Aaron~Grattafiori et~al. 2024{\natexlab{a}}.
\newblock \href {https://arxiv.org/abs/2407.21783} {The llama 3 herd of models}.
\newblock \emph{Preprint}, arXiv:2407.21783.

\bibitem[{et~al.(2023{\natexlab{a}})}]{jiang2023mistral7b}
Albert Q.~Jiang et~al. 2023{\natexlab{a}}.
\newblock \href {https://arxiv.org/abs/2310.06825} {Mistral 7b}.
\newblock \emph{Preprint}, arXiv:2310.06825.

\bibitem[{et~al.(2021)}]{hu2021loralowrankadaptationlarge}
Edward J.~Hu et~al. 2021.
\newblock \href {https://arxiv.org/abs/2106.09685} {Lora: Low-rank adaptation of large language models}.
\newblock \emph{Preprint}, arXiv:2106.09685.

\bibitem[{et~al.(2019)}]{bredin2019pyannoteaudioneuralbuildingblocks}
Hervé~Bredin et~al. 2019.
\newblock \href {https://arxiv.org/abs/1911.01255} {pyannote.audio: neural building blocks for speaker diarization}.
\newblock \emph{Preprint}, arXiv:1911.01255.

\bibitem[{et~al.(2023{\natexlab{b}})}]{bai2023qwentechnicalreport}
Jinze~Bai et~al. 2023{\natexlab{b}}.
\newblock \href {https://arxiv.org/abs/2309.16609} {Qwen technical report}.
\newblock \emph{Preprint}, arXiv:2309.16609.

\bibitem[{et~al.(2024{\natexlab{b}})}]{openai2024gpt4ocard}
OpenAI et~al. 2024{\natexlab{b}}.
\newblock \href {https://arxiv.org/abs/2410.21276} {Gpt-4o system card}.
\newblock \emph{Preprint}, arXiv:2410.21276.

\bibitem[{et~al.(2023{\natexlab{c}})}]{ma2023emotion2vecselfsupervisedpretrainingspeech}
Ziyang~Ma et~al. 2023{\natexlab{c}}.
\newblock \href {https://arxiv.org/abs/2312.15185} {emotion2vec: Self-supervised pre-training for speech emotion representation}.
\newblock \emph{Preprint}, arXiv:2312.15185.

\bibitem[{{FFmpeg Developers}(2023)}]{ffmpeg}
{FFmpeg Developers}. 2023.
\newblock {FFmpeg tool (Version 4.4.1)}.
\newblock {Software available from \url{http://ffmpeg.org/}}.
\newblock Accessed: 2023-10-05.

\bibitem[{Fu et~al.(2024{\natexlab{a}})Fu, Dai, Luo, Li, Ren, Zhang, Wang, Zhou, Shen, Zhang, Chen, Li, Lin, Zhao, Li, Xu, Zheng, Chen, Ji, and Sun}]{fu2024videommefirstevercomprehensiveevaluation}
Chaoyou Fu, Yuhan Dai, Yongdong Luo, Lei Li, Shuhuai Ren, Renrui Zhang, Zihan Wang, Chenyu Zhou, Yunhang Shen, Mengdan Zhang, Peixian Chen, Yanwei Li, Shaohui Lin, Sirui Zhao, Ke~Li, Tong Xu, Xiawu Zheng, Enhong Chen, Rongrong Ji, and Xing Sun. 2024{\natexlab{a}}.
\newblock \href {https://arxiv.org/abs/2405.21075} {Video-mme: The first-ever comprehensive evaluation benchmark of multi-modal llms in video analysis}.
\newblock \emph{Preprint}, arXiv:2405.21075.

\bibitem[{Fu et~al.(2024{\natexlab{b}})Fu, Lin, Long, Shen, Zhao, Zhang, Dong, Wang, Yin, Ma, Zheng, He, Ji, Wu, Shan, and Sun}]{fu2024vitaopensourceinteractiveomni}
Chaoyou Fu, Haojia Lin, Zuwei Long, Yunhang Shen, Meng Zhao, Yifan Zhang, Shaoqi Dong, Xiong Wang, Di~Yin, Long Ma, Xiawu Zheng, Ran He, Rongrong Ji, Yunsheng Wu, Caifeng Shan, and Xing Sun. 2024{\natexlab{b}}.
\newblock \href {https://arxiv.org/abs/2408.05211} {Vita: Towards open-source interactive omni multimodal llm}.
\newblock \emph{Preprint}, arXiv:2408.05211.

\bibitem[{Girdhar et~al.(2023)Girdhar, El-Nouby, Liu, Singh, Alwala, Joulin, and Misra}]{girdhar2023imagebindembeddingspacebind}
Rohit Girdhar, Alaaeldin El-Nouby, Zhuang Liu, Mannat Singh, Kalyan~Vasudev Alwala, Armand Joulin, and Ishan Misra. 2023.
\newblock \href {https://arxiv.org/abs/2305.05665} {Imagebind: One embedding space to bind them all}.
\newblock \emph{Preprint}, arXiv:2305.05665.

\bibitem[{Goyal et~al.(2017)Goyal, Kahou, Michalski, Materzyńska, Westphal, Kim, Haenel, Fruend, Yianilos, Mueller-Freitag, Hoppe, Thurau, Bax, and Memisevic}]{goyal2017somethingsomethingvideodatabase}
Raghav Goyal, Samira~Ebrahimi Kahou, Vincent Michalski, Joanna Materzyńska, Susanne Westphal, Heuna Kim, Valentin Haenel, Ingo Fruend, Peter Yianilos, Moritz Mueller-Freitag, Florian Hoppe, Christian Thurau, Ingo Bax, and Roland Memisevic. 2017.
\newblock \href {https://arxiv.org/abs/1706.04261} {The "something something" video database for learning and evaluating visual common sense}.
\newblock \emph{Preprint}, arXiv:1706.04261.

\bibitem[{Huang et~al.(2023)Huang, Dong, Wang, Hao, Singhal, Ma, Lv, Cui, Mohammed, Patra, Liu, Aggarwal, Chi, Bjorck, Chaudhary, Som, Song, and Wei}]{huang2023languageneedaligningperception}
Shaohan Huang, Li~Dong, Wenhui Wang, Yaru Hao, Saksham Singhal, Shuming Ma, Tengchao Lv, Lei Cui, Owais~Khan Mohammed, Barun Patra, Qiang Liu, Kriti Aggarwal, Zewen Chi, Johan Bjorck, Vishrav Chaudhary, Subhojit Som, Xia Song, and Furu Wei. 2023.
\newblock \href {https://arxiv.org/abs/2302.14045} {Language is not all you need: Aligning perception with language models}.
\newblock \emph{Preprint}, arXiv:2302.14045.

\bibitem[{Kay et~al.(2017)Kay, Carreira, Simonyan, Zhang, Hillier, Vijayanarasimhan, Viola, Green, Back, Natsev, Suleyman, and Zisserman}]{kay2017kineticshumanactionvideo}
Will Kay, Joao Carreira, Karen Simonyan, Brian Zhang, Chloe Hillier, Sudheendra Vijayanarasimhan, Fabio Viola, Tim Green, Trevor Back, Paul Natsev, Mustafa Suleyman, and Andrew Zisserman. 2017.
\newblock \href {https://arxiv.org/abs/1705.06950} {The kinetics human action video dataset}.
\newblock \emph{Preprint}, arXiv:1705.06950.

\bibitem[{Kheddar et~al.(2024)Kheddar, Hemis, and Himeur}]{Kheddar_2024}
Hamza Kheddar, Mustapha Hemis, and Yassine Himeur. 2024.
\newblock \href {https://doi.org/10.1016/j.inffus.2024.102422} {Automatic speech recognition using advanced deep learning approaches: A survey}.
\newblock \emph{Information Fusion}, 109:102422.

\bibitem[{Lee et~al.(2024)Lee, Wang, Fan, Zhang, Liu, Hao, Bhat, and Li}]{lee2024nowyousee}
Seon-Ho Lee, Jue Wang, David Fan, Zhikang Zhang, Linda Liu, Xiang Hao, Vimal Bhat, and Xinyu Li. 2024.
\newblock \href {https://arxiv.org/abs/2412.10002} {Nowyousee me: Context-aware automatic audio description}.
\newblock \emph{Preprint}, arXiv:2412.10002.

\bibitem[{Li et~al.(2023)Li, Wang, Wang, Ge, Ge, and Shan}]{li2023seedbenchbenchmarkingmultimodalllms}
Bohao Li, Rui Wang, Guangzhi Wang, Yuying Ge, Yixiao Ge, and Ying Shan. 2023.
\newblock \href {https://arxiv.org/abs/2307.16125} {Seed-bench: Benchmarking multimodal llms with generative comprehension}.
\newblock \emph{Preprint}, arXiv:2307.16125.

\bibitem[{Li et~al.(2020)Li, Wong, Zhao, and Kankanhalli}]{Li_2020}
Junnan Li, Yongkang Wong, Qi~Zhao, and Mohan~S. Kankanhalli. 2020.
\newblock \href {https://doi.org/10.1109/tmm.2019.2930041} {Video storytelling: Textual summaries for events}.
\newblock \emph{IEEE Transactions on Multimedia}, 22(2):554–565.

\bibitem[{Li et~al.(2024{\natexlab{a}})Li, He, Wang, Li, Wang, Luo, Wang, Wang, and Qiao}]{li2024videochatchatcentricvideounderstanding}
KunChang Li, Yinan He, Yi~Wang, Yizhuo Li, Wenhai Wang, Ping Luo, Yali Wang, Limin Wang, and Yu~Qiao. 2024{\natexlab{a}}.
\newblock \href {https://arxiv.org/abs/2305.06355} {Videochat: Chat-centric video understanding}.
\newblock \emph{Preprint}, arXiv:2305.06355.

\bibitem[{Li et~al.(2024{\natexlab{b}})Li, Wang, He, Li, Wang, Liu, Wang, Xu, Chen, Luo, Wang, and Qiao}]{li2024mvbenchcomprehensivemultimodalvideo}
Kunchang Li, Yali Wang, Yinan He, Yizhuo Li, Yi~Wang, Yi~Liu, Zun Wang, Jilan Xu, Guo Chen, Ping Luo, Limin Wang, and Yu~Qiao. 2024{\natexlab{b}}.
\newblock \href {https://arxiv.org/abs/2311.17005} {Mvbench: A comprehensive multi-modal video understanding benchmark}.
\newblock \emph{Preprint}, arXiv:2311.17005.

\bibitem[{Lin et~al.(2024)Lin, Ye, Zhu, Cui, Ning, Jin, and Yuan}]{lin2024videollavalearningunitedvisual}
Bin Lin, Yang Ye, Bin Zhu, Jiaxi Cui, Munan Ning, Peng Jin, and Li~Yuan. 2024.
\newblock \href {https://arxiv.org/abs/2311.10122} {Video-llava: Learning united visual representation by alignment before projection}.
\newblock \emph{Preprint}, arXiv:2311.10122.

\bibitem[{Lin(2004)}]{Lin2004ROUGE}
Chin-Yew Lin. 2004.
\newblock \href {https://aclanthology.org/W04-1013/} {Rouge: A package for automatic evaluation of summaries}.
\newblock In \emph{Text Summarization Branches Out}, pages 74--81, Barcelona, Spain. Association for Computational Linguistics.

\bibitem[{Liu et~al.(2023)Liu, Li, Wu, and Lee}]{liu2023visualinstructiontuning}
Haotian Liu, Chunyuan Li, Qingyang Wu, and Yong~Jae Lee. 2023.
\newblock \href {https://arxiv.org/abs/2304.08485} {Visual instruction tuning}.
\newblock \emph{Preprint}, arXiv:2304.08485.

\bibitem[{Liu et~al.(2024)Liu, Duan, Zhang, Li, Zhang, Zhao, Yuan, Wang, He, Liu, Chen, and Lin}]{liu2024mmbenchmultimodalmodelallaround}
Yuan Liu, Haodong Duan, Yuanhan Zhang, Bo~Li, Songyang Zhang, Wangbo Zhao, Yike Yuan, Jiaqi Wang, Conghui He, Ziwei Liu, Kai Chen, and Dahua Lin. 2024.
\newblock \href {https://arxiv.org/abs/2307.06281} {Mmbench: Is your multi-modal model an all-around player?}
\newblock \emph{Preprint}, arXiv:2307.06281.

\bibitem[{Lyu et~al.(2023)Lyu, Wu, Wang, Huang, Liu, Du, Shi, and Tu}]{lyu2023macawllm}
Chenyang Lyu, Minghao Wu, Longyue Wang, Xinting Huang, Bingshuai Liu, Zefeng Du, Shuming Shi, and Zhaopeng Tu. 2023.
\newblock \href {https://arxiv.org/abs/2306.09093} {Macaw-llm: Multi-modal language modeling with image, audio, video, and text integration}.
\newblock \emph{Preprint}, arXiv:2306.09093.

\bibitem[{Maaz et~al.(2024)Maaz, Rasheed, Khan, and Khan}]{maaz2024videochatgptdetailedvideounderstanding}
Muhammad Maaz, Hanoona Rasheed, Salman Khan, and Fahad~Shahbaz Khan. 2024.
\newblock \href {https://arxiv.org/abs/2306.05424} {Video-chatgpt: Towards detailed video understanding via large vision and language models}.
\newblock \emph{Preprint}, arXiv:2306.05424.

\bibitem[{OpenAI(2022)}]{openai2022whisper}
OpenAI. 2022.
\newblock Whisper: A general-purpose speech recognition model.
\newblock \url{https://cdn.openai.com/papers/whisper.pdf}.
\newblock Accessed: 2023-10-06.

\bibitem[{Papineni et~al.(2002)Papineni, Roukos, Ward, and Zhu}]{Papineni2002BLEU}
Kishore Papineni, Salim Roukos, Todd Ward, and Wei-Jing Zhu. 2002.
\newblock \href {https://doi.org/10.3115/1073083.1073135} {Bleu: a method for automatic evaluation of machine translation}.
\newblock In \emph{Proceedings of the 40th Annual Meeting of the Association for Computational Linguistics}, pages 311--318. Association for Computational Linguistics.

\bibitem[{Pătrăucean et~al.(2023)Pătrăucean, Smaira, Gupta, Continente, Markeeva, Banarse, Koppula, Heyward, Malinowski, Yang, Doersch, Matejovicova, Sulsky, Miech, Frechette, Klimczak, Koster, Zhang, Winkler, Aytar, Osindero, Damen, Zisserman, and Carreira}]{pătrăucean2023perceptiontestdiagnosticbenchmark}
Viorica Pătrăucean, Lucas Smaira, Ankush Gupta, Adrià~Recasens Continente, Larisa Markeeva, Dylan Banarse, Skanda Koppula, Joseph Heyward, Mateusz Malinowski, Yi~Yang, Carl Doersch, Tatiana Matejovicova, Yury Sulsky, Antoine Miech, Alex Frechette, Hanna Klimczak, Raphael Koster, Junlin Zhang, Stephanie Winkler, Yusuf Aytar, Simon Osindero, Dima Damen, Andrew Zisserman, and João Carreira. 2023.
\newblock \href {https://arxiv.org/abs/2305.13786} {Perception test: A diagnostic benchmark for multimodal video models}.
\newblock \emph{Preprint}, arXiv:2305.13786.

\bibitem[{Radford et~al.(2022)Radford, Kim, Xu, Brockman, McLeavey, and Sutskever}]{radford2022robustspeechrecognitionlargescale}
Alec Radford, Jong~Wook Kim, Tao Xu, Greg Brockman, Christine McLeavey, and Ilya Sutskever. 2022.
\newblock \href {https://arxiv.org/abs/2212.04356} {Robust speech recognition via large-scale weak supervision}.
\newblock \emph{Preprint}, arXiv:2212.04356.

\bibitem[{Schroff et~al.(2015)Schroff, Kalenichenko, and Philbin}]{face}
Florian Schroff, Dmitry Kalenichenko, and James Philbin. 2015.
\newblock \href {https://doi.org/10.1109/cvpr.2015.7298682} {Facenet: A unified embedding for face recognition and clustering}.
\newblock In \emph{2015 IEEE Conference on Computer Vision and Pattern Recognition (CVPR)}, page 815–823. IEEE.

\bibitem[{Touvron et~al.(2023)Touvron, Lavril, Izacard, Martinet, Lachaux, Lacroix, Rozière, Goyal, Hambro, Azhar, Rodriguez, Joulin, Grave, and Lample}]{touvron2023llamaopenefficientfoundation}
Hugo Touvron, Thibaut Lavril, Gautier Izacard, Xavier Martinet, Marie-Anne Lachaux, Timothée Lacroix, Baptiste Rozière, Naman Goyal, Eric Hambro, Faisal Azhar, Aurelien Rodriguez, Armand Joulin, Edouard Grave, and Guillaume Lample. 2023.
\newblock \href {https://arxiv.org/abs/2302.13971} {Llama: Open and efficient foundation language models}.
\newblock \emph{Preprint}, arXiv:2302.13971.

\bibitem[{Tseng et~al.(2024)Tseng, Berry, Chen, Chiu, Lin, Liu, Peng, Shih, Wang, Wu, Huang, Lai, Li, Harwath, Tsao, Watanabe, Mohamed, Feng, and yi~Lee}]{tseng2024avsuperbmultitaskevaluationbenchmark}
Yuan Tseng, Layne Berry, Yi-Ting Chen, I-Hsiang Chiu, Hsuan-Hao Lin, Max Liu, Puyuan Peng, Yi-Jen Shih, Hung-Yu Wang, Haibin Wu, Po-Yao Huang, Chun-Mao Lai, Shang-Wen Li, David Harwath, Yu~Tsao, Shinji Watanabe, Abdelrahman Mohamed, Chi-Luen Feng, and Hung yi~Lee. 2024.
\newblock \href {https://arxiv.org/abs/2309.10787} {Av-superb: A multi-task evaluation benchmark for audio-visual representation models}.
\newblock \emph{Preprint}, arXiv:2309.10787.

\bibitem[{Wang et~al.(2024)Wang, Li, Li, Yu, He, Wang, Chen, Pei, Yan, Zheng, Xu, Wang, Shi, Jiang, Li, Zhang, Huang, Qiao, Wang, and Wang}]{wang2024internvideo2scalingfoundationmodels}
Yi~Wang, Kunchang Li, Xinhao Li, Jiashuo Yu, Yinan He, Chenting Wang, Guo Chen, Baoqi Pei, Ziang Yan, Rongkun Zheng, Jilan Xu, Zun Wang, Yansong Shi, Tianxiang Jiang, Songze Li, Hongjie Zhang, Yifei Huang, Yu~Qiao, Yali Wang, and Limin Wang. 2024.
\newblock \href {https://arxiv.org/abs/2403.15377} {Internvideo2: Scaling foundation models for multimodal video understanding}.
\newblock \emph{Preprint}, arXiv:2403.15377.

\bibitem[{Wang et~al.(2022)Wang, Li, Li, He, Huang, Zhao, Zhang, Xu, Liu, Wang, Xing, Chen, Pan, Yu, Wang, Wang, and Qiao}]{wang2022internvideogeneralvideofoundation}
Yi~Wang, Kunchang Li, Yizhuo Li, Yinan He, Bingkun Huang, Zhiyu Zhao, Hongjie Zhang, Jilan Xu, Yi~Liu, Zun Wang, Sen Xing, Guo Chen, Junting Pan, Jiashuo Yu, Yali Wang, Limin Wang, and Yu~Qiao. 2022.
\newblock \href {https://arxiv.org/abs/2212.03191} {Internvideo: General video foundation models via generative and discriminative learning}.
\newblock \emph{Preprint}, arXiv:2212.03191.

\bibitem[{Xiao et~al.(2021)Xiao, Shang, Yao, and Chua}]{xiao2021nextqanextphasequestionansweringexplaining}
Junbin Xiao, Xindi Shang, Angela Yao, and Tat-Seng Chua. 2021.
\newblock \href {https://arxiv.org/abs/2105.08276} {Next-qa:next phase of question-answering to explaining temporal actions}.
\newblock \emph{Preprint}, arXiv:2105.08276.

\bibitem[{Xie et~al.(2024)Xie, Zhang, Zhou, Li, Zhang, Hessel, Yang, and Liu}]{xie2024funqasurprisingvideocomprehension}
Binzhu Xie, Sicheng Zhang, Zitang Zhou, Bo~Li, Yuanhan Zhang, Jack Hessel, Jingkang Yang, and Ziwei Liu. 2024.
\newblock \href {https://arxiv.org/abs/2306.14899} {Funqa: Towards surprising video comprehension}.
\newblock \emph{Preprint}, arXiv:2306.14899.

\bibitem[{Xu et~al.(2017)Xu, Zhao, Xiao, Wu, Zhang, He, and Zhuang}]{xu2017video}
Dejing Xu, Zhou Zhao, Jun Xiao, Fei Wu, Hanwang Zhang, Xiangnan He, and Yueting Zhuang. 2017.
\newblock Video question answering via gradually refined attention over appearance and motion.
\newblock In \emph{ACM Multimedia}.

\bibitem[{Xu et~al.(2016)Xu, Mei, Yao, and Rui}]{7780940}
Jun Xu, Tao Mei, Ting Yao, and Yong Rui. 2016.
\newblock \href {https://doi.org/10.1109/CVPR.2016.571} {Msr-vtt: A large video description dataset for bridging video and language}.
\newblock In \emph{2016 IEEE Conference on Computer Vision and Pattern Recognition (CVPR)}, pages 5288--5296.

\bibitem[{Xu et~al.(2023)Xu, Shao, Zhang, Gao, Liu, Lei, Meng, Huang, Qiao, and Luo}]{xu2023lvlmehubcomprehensiveevaluationbenchmark}
Peng Xu, Wenqi Shao, Kaipeng Zhang, Peng Gao, Shuo Liu, Meng Lei, Fanqing Meng, Siyuan Huang, Yu~Qiao, and Ping Luo. 2023.
\newblock \href {https://arxiv.org/abs/2306.09265} {Lvlm-ehub: A comprehensive evaluation benchmark for large vision-language models}.
\newblock \emph{Preprint}, arXiv:2306.09265.

\bibitem[{Yang et~al.(2024)Yang, Zhan, Wang, Wang, Ge, Zheng, and Jin}]{yang2024synchronizedvideostorytellinggenerating}
Dingyi Yang, Chunru Zhan, Ziheng Wang, Biao Wang, Tiezheng Ge, Bo~Zheng, and Qin Jin. 2024.
\newblock \href {https://arxiv.org/abs/2405.14040} {Synchronized video storytelling: Generating video narrations with structured storyline}.
\newblock \emph{Preprint}, arXiv:2405.14040.

\bibitem[{Yaseen(2024)}]{yaseen2024yolov8indepthexplorationinternal}
Muhammad Yaseen. 2024.
\newblock \href {https://arxiv.org/abs/2408.15857} {What is yolov8: An in-depth exploration of the internal features of the next-generation object detector}.
\newblock \emph{Preprint}, arXiv:2408.15857.

\bibitem[{Ye et~al.(2024{\natexlab{a}})Ye, Xu, Xu, Ye, Yan, Zhou, Wang, Hu, Shi, Shi, Li, Xu, Chen, Tian, Qian, Zhang, Huang, and Zhou}]{ye2024mplugowlmodularizationempowerslarge}
Qinghao Ye, Haiyang Xu, Guohai Xu, Jiabo Ye, Ming Yan, Yiyang Zhou, Junyang Wang, Anwen Hu, Pengcheng Shi, Yaya Shi, Chenliang Li, Yuanhong Xu, Hehong Chen, Junfeng Tian, Qi~Qian, Ji~Zhang, Fei Huang, and Jingren Zhou. 2024{\natexlab{a}}.
\newblock \href {https://arxiv.org/abs/2304.14178} {mplug-owl: Modularization empowers large language models with multimodality}.
\newblock \emph{Preprint}, arXiv:2304.14178.

\bibitem[{Ye et~al.(2024{\natexlab{b}})Ye, Chen, Li, Xin, Li, Zhou, and Bu}]{ye-etal-2024-mmad}
Xiaojun Ye, Junhao Chen, Xiang Li, Haidong Xin, Chao Li, Sheng Zhou, and Jiajun Bu. 2024{\natexlab{b}}.
\newblock \href {https://aclanthology.org/2024.lrec-main.998/} {{MMAD}:multi-modal movie audio description}.
\newblock In \emph{Proceedings of the 2024 Joint International Conference on Computational Linguistics, Language Resources and Evaluation (LREC-COLING 2024)}, pages 11415--11428, Torino, Italia. ELRA and ICCL.

\bibitem[{Yu et~al.(2024)Yu, Yang, Li, Wang, Lin, Liu, Wang, and Wang}]{yu2024mmvetevaluatinglargemultimodal}
Weihao Yu, Zhengyuan Yang, Linjie Li, Jianfeng Wang, Kevin Lin, Zicheng Liu, Xinchao Wang, and Lijuan Wang. 2024.
\newblock \href {https://arxiv.org/abs/2308.02490} {Mm-vet: Evaluating large multimodal models for integrated capabilities}.
\newblock \emph{Preprint}, arXiv:2308.02490.

\bibitem[{Zhao et~al.(2024)Zhao, Zhao, Zhou, Wu, Chu, Miao, Schroff, Adam, Liu, Gong, Krähenbühl, and Yuan}]{zhao2024distillingvisionlanguagemodelsmillions}
Yue Zhao, Long Zhao, Xingyi Zhou, Jialin Wu, Chun-Te Chu, Hui Miao, Florian Schroff, Hartwig Adam, Ting Liu, Boqing Gong, Philipp Krähenbühl, and Liangzhe Yuan. 2024.
\newblock \href {https://arxiv.org/abs/2401.06129} {Distilling vision-language models on millions of videos}.
\newblock \emph{Preprint}, arXiv:2401.06129.

\bibitem[{Zhu et~al.(2023)Zhu, Chen, Shen, Li, and Elhoseiny}]{zhu2023minigpt4enhancingvisionlanguageunderstanding}
Deyao Zhu, Jun Chen, Xiaoqian Shen, Xiang Li, and Mohamed Elhoseiny. 2023.
\newblock \href {https://arxiv.org/abs/2304.10592} {Minigpt-4: Enhancing vision-language understanding with advanced large language models}.
\newblock \emph{Preprint}, arXiv:2304.10592.

\end{thebibliography}
